\documentclass[10pt,twocolumn,letterpaper]{article}
\usepackage{3dv}
\usepackage{times}
\usepackage{epsfig}
\usepackage{graphicx}
\usepackage{amsmath}
\usepackage{amssymb}
 \usepackage{cite}

\usepackage[pagebackref=true,breaklinks=true,letterpaper=true,colorlinks,bookmarks=false]{hyperref}

\threedvfinalcopy 

\def\threedvPaperID{0061} 

\ifthreedvfinal\fi
\DeclareMathOperator*{\argmin}{arg\,min}
\renewcommand*{\arraystretch}{.6}
\usepackage{tablefootnote}
\usepackage{booktabs}
\DeclareMathOperator*{\argmax}{arg\,max}
\newcommand{\rec}{Q}
\graphicspath{{Images/}}
\DeclareGraphicsExtensions{.png,.pdf,.jpg,.mps,.jpeg,.jbig2,.jb2,.JPG,.JPEG,.JBIG2,.JB2}

\begin{document}

\title{Rethinking Pose in 3D: Multi-stage Refinement and Recovery for Markerless Motion Capture}


\author{Denis Tome\\
UCL\\
\and
Matteo Toso\\
University of Surrey\\
\and
Lourdes Agapito\\
UCL
\and
Chris Russell\\
University of Surrey
}

\maketitle
\thispagestyle{empty}
\begin{abstract}
 We propose a CNN-based approach for multi-camera markerless motion capture of
 the human body. Unlike existing methods that first perform pose estimation
 on individual cameras and generate 3D models as post-processing, 
 our approach makes use of 3D reasoning throughout a multi-stage approach.
 This novelty
 allows us to use provisional 3D models of human pose to rethink where the joints
 should be located in the image and to recover from past mistakes. Our
 principled refinement of 3D human poses lets us make use of image cues, even
 from images where we previously misdetected joints, to refine our estimates
 as part of an end-to-end approach. Finally, we demonstrate how the high-quality output
 of our multi-camera setup can be used as an additional training source to
 improve the accuracy of existing single camera models.
%
\end{abstract}

\section{Introduction}

One fundamental challenge in the 3D estimation of dynamic and moving objects
lies in finding a rich source of ground-truth data. This is not just a problem
for modern learning based approaches, that require an abundance of data in order
to make inferences about the world, but also for the traditional ones such
as model-based reasoning that make heavy use of constraining prior
information about the world. Even these traditional methods rely on carefully
tuned parameters which control expressiveness of the model~\cite{bregler2000recovering}, internal connectivity
priors~\cite{russell2014video}, or both~\cite{garg2013dense} that must be
adjusted to recover plausible reconstructions.

Extracting 3D data from images is a fundamentally ill-posed problem that even
people find challenging. Unlike standard image labelling problems, such as
Imagenet~\cite{deng2009imagenet}, that make heavy use of human annotation, we
cannot simply expect people to reliably annotate images with the distance of
joints from the camera. The gold standard for accurately capturing 3D
information of full-body human poses data remains using Multi-camera Motion Capture
(MoCap) systems. These systems make use of early vision techniques based on the
identification of markers across multiple cameras and on the estimation of the 3D
location of these points through triangulation. However, these systems require
strong, unambiguous cues to identify the points. In practice, this means that
successful MoCap relies on the subject wearing dark tight clothing and
brightly coloured markers, making the captured images unrepresentative of
the natural scenes we wish to reconstruct.

In response to these limitations, some recent
works~\cite{ionescu2014human3,rogez2016MoCap,varol2017learning} have generated
more varied synthetic images using MoCap pose data as the source of the human
poses. Although these images are more varied than MoCap data, they are still not
natural images; and these images tend not to capture information and confusion
caused by the deformation of loose fitting clothing~\cite{lassner2017generative}.

Another approach to avoiding these problems is to chain together different
regressors based on multiple data sources; one network is trained to
predict 2D joint locations in natural images, while a second regressor upgrades
these 2D joint locations to 3D using MoCap data. This approach comes with
caveats similar to those of the methods discussed above. We might know that a
method gives highly accurate 3D poses on MoCap data and good 2D joint locations
in natural images, but we remain fundamentally unsure as to its 3D accuracy in
natural images.

\begin{figure*}[!ht]
  \begin{center}
    \includegraphics[width=1.8\columnwidth]{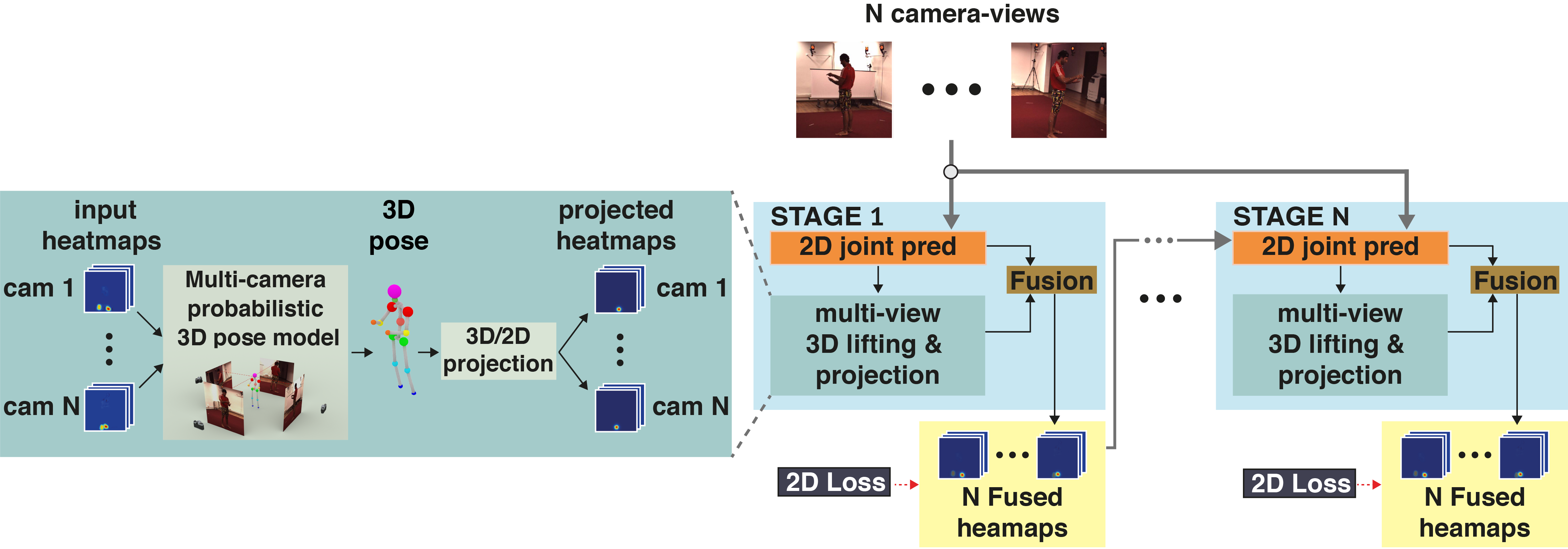}
  \end{center}
  \caption{Multi-stage architecture of our proposed multi-camera 3D
   human pose network. Each stage takes  images
   from all the camera views and the set of per-image 2D joints
   (expressed as heatmaps) predicted in the previous stage and outputs
   a refined prediction. In each stage, the 2D predictions from all
   views are used to reconstruct a single 3D pose, consistent with all
   camera views. This 3D pose is projected back into the image and used to
   improve predictions in the next stage. See section \ref{sec:network} for more
   details.\label{fig:architecture_stage}}
\end{figure*}

%
%
As such, effective markerless motion capture is an important tool to train
networks to generate reliable 3D models from natural images. We present a Huber
loss based robust
estimator 
for fusing multi-view 2D pose predictions into a coherent 3D pose,
consistent with natural human poses. Unlike existing 3D frameworks, this is not
simply done at the end of a pipeline for 2D joint estimation, but is iterated
through multiple-stages. This carries substantial benefits. Our use of a robust
estimator means that at each stage the 3D model can discard a minority of
incorrect 2D joint estimates; the knowledge of where the joints should
be in each image is fed back into the algorithm for image-based refinement.



One fundamental question regarding these datasets composed of millions of frames,
such as Human3.6M, is whether they are in fact large enough. The primary
issue is whether the dataset is sufficiently diverse to allow trained networks
to exhibit good generalisation to a held-out test set. Even in restrictive
cases, such as the test set used in Human3.6M, where the held-out data consists
of new actors performing the same movements in similar clothes in the same
studio, there is enough variability in individual body shapes and in how they
move that generalisation is not guaranteed.

To help address this issue, we demonstrate how unlabelled data can be
labelled by our algorithm and augment the datasets used for the
training of existing methods, leading to overall better performance on
standard benchmarks. We evaluate multiple networks and find consistent
multi-millimetre improvement.  When the differences between
state-of-the-art networks are so small, this raises questions as to
whether we are over-fitting and if time would be better spent building
larger datasets rather than fighting for small improvements obtained
from architectural changes.

{\textbf{Our contribution:} We extended existing work on single view
  reconstruction to a multi-camera setting and show how such single
  view methods can be enhanced by training on multiview based
  annotation of unlabelled data.  Our use of an iterative, and robust,
  multistage approach to multi-view reconstruction allows us to
  correct mistakes in body joint estimations as they arise, and to {\em
    think again}, reconsidering the 2D position of joints in the image
  using interim knowledge of 3D pose.
  
  Unlike the multiview bootstrapping of Simon \etal~\cite{mvbootstrapping} which
  iteratively retrains 2D estimators, our refinement happens at test-time, not
  training, and only makes use of the information contained in a single set of
  images captured at one moment in time, rather than requiring extensive
  retraining on a larger dataset. As such, our approach can be seen as complementary
  to theirs.
}
\section{Prior Work}

Deep convolutional neural networks have led to a substantial
improvement in 3D human pose estimation from one or several
images. This task is challenging as it involves solving two ill-posed
problems: correctly localising the joints of the human body within 2D
images and correctly lifting them in 3D. The 2D visual recognition
task of localising body joints in the image is made difficult by
multiple possible confusing factors including occlusion, variability
in the colour, shape and texture of clothing and the lighting
conditions, while the task of lifting into 3D is even challenging for
humans and intrinsically limited by the existence of perspective
ambiguities.


We now review the four most dominant paradigms in monocular 3D human pose
estimation: \emph{(i)} direct image to 3D pose regression; \emph{(ii)}
3D pose estimation from 2D joint estimates; \emph{(iii)} joint 2D and
3D pose estimation; and \emph{(iv)} 3D pose estimation trained on 2D
reprojection loss. We  also cover recent deep-learning based approaches to
multi-view 3D pose estimation. 

{\bf Direct human 3D pose from a single image:}
Many recent approaches treat 3D pose estimation from a single input image as a
fully supervised learning problem and make use of deep architectures to directly
regress the 3D coordinates of human joints from the image
~\cite{li20143D,park20163D,tekin2016structured,zhou2016deep}. Much of the
novelty of more recent works has involved combining end-to-end learning with
expressive 3D priors to constrain the final 3D pose. Li and Chan~\cite{li20143D}
proposed strategies to jointly train for pose regression and body part
detection, Tekin \etal~\cite{tekin2016structured} used a pre-trained
auto-encoder to enforce structural constraints on the output skeleton. Li
\etal~\cite{li2015maximum} trained a deep neural network to predict
similarity scores between an input image and a 3D pose using a max-margin loss.
Zhou \etal~\cite{zhou2016deep} enforce bone lengths in predictions. Tekin
\etal also leverage 2D image data~\cite{tekin2017learning} by adding a
second network stream whose outputs are fused with the 3D regressor. Following
the trend in 2D human pose estimation to predict heatmaps rather than regressing
2D landmarks, Pavlakos~\cite{pavlakos2017coarse} predicted per-voxel likelihoods,
or 3D heatmaps, for each joint using a coarse-to-fine approach.

These methods share the disadvantage of generalising poorly to images in the
wild: the need for ground truth 3D poses to train the image to 3D pose
regressor means that they must be trained exclusively on images captured in
MoCap studios, with all the limitations that come with it.

{\bf 3D pose from 2D joint estimates:} The recent success of 2D pose detection
has led to a proliferation of two-stage approaches that estimate 3D human poses
from 2D landmarks. Detections are obtained from off-the-shelf
2D pose detectors such
as~\cite{wei2016convolutional,newell2016stacked,pishchulin2016deepcut}; or
included as an initial step in the estimation~\cite{lin2017recurrent}. The task
is then to lift the 2D coordinates into 3D either by model
fitting~\cite{ramakrishna2012reconstructing,akhter2015pose,zhou2017sparse,zhou2016sparseness,bogo2016keep,sanzari2016bayesian}
or regression~\cite{martinez2017simple,moreno20173d}.
Moreno-Noguer~\cite{moreno20173d} estimated 3D pose from 2D inputs using
2D-to-3D distance matrix regression. Chen and Ramanan~\cite{chen20173d}
estimated the depth of 2D landmarks by matching them to a library of 3D poses.
Bogo \etal~\cite{bogo2016keep} fitted a dense statistical shape and pose model,
trained on thousands of 3D scans~\cite{loper2015smpl}, to 2D joints obtained
with DeepCut~\cite{pishchulin2016deepcut}; while Sanzari \emph{et
  al.}~\cite{sanzari2016bayesian} fitted a non-parametric probabilistic pose
model. Martinez \etal~\cite{martinez2017simple} show how even a simple regressor
- a feed-forward network with residual connections and batch normalization -
vastly outperforms previous approaches when given ground truth 2D landmarks as
input, suggesting that the largest source of errors in 3D pose reconstruction is
incorrect 2D estimation.

{\bf Joint 2D-3D pose estimation:} 
Several monocular approaches solve for 2D and
3D pose
jointly~\cite{simo2013joint,tekin2017learning,rogez2017lcr,sun2017compositional}.
Rogez \etal~\cite{rogez2017lcr} proposed an end-to-end
architecture that combines a region proposal network for human
localisation with classification and regression branches for joint
estimation of 2D and 3D human pose. Sun \emph{et
 al.}~\cite{sun2017compositional} adopted a bone based representation
for the pose and propose a unified setting for 2D and 3D pose
estimation that encoded long range interactions between bones. Both
approaches achieve best results when a 2D loss is combined with the
standard 3D loss. Zhou \etal~\cite{zhou2017towards} shared
common representations between the 2D and the 3D tasks inside the
network which is trained end-to-end with both 2D and 3D losses. 

{\bf Training with 2D-only loss:} A few recent approaches bypass the need to
annotate images with 3D ground truth labels by keeping an internal 3D
representation of the pose but training based on 2D reprojection losses. These
approaches benefit from both their ability to generalise to in-the-wild images as they
do not rely on 3D annotated images that can only be captured in studios; and the
added structural 3D pose priors afforded by internal 3D representation. Tome
\etal~\cite{tome2017lifting} proposed a multi-stage architecture that reasons
jointly about 2D and 3D pose to improve both tasks. Key to their architecture is
a 3D lifting module that reconstructs 2D estimated landmarks in 3D and projects
them back into 2D, as their end-to-end training minimises deviations of the
reprojected 3D landmarks from the ground truth 2D labels. Wu \etal's single
image 3D interpreter network~\cite{wu2016single} also uses a loss based on the
2D re-projection error of predicted 3D landmarks, along with a supervised 2D
landmarks to 3D pose regressor. Tung \etal~\cite{tung2017adversarial} combine a
similar 2D reprojection loss with an adversarial loss and
later~\cite{tung2017self} propose to combine strong supervision from synthetic
data with a self-supervised loss based on consistency checks against 2D
estimates of keypoints, segmentation and optical flow.

{\bf Multi-view human pose:} Elhayek \etal~\cite{elhayek2015efficient}
fused 2D body part detections, from a ConvNet-based 2D pose estimation,
with a generative model-based multi-view tracking algorithm to reconstruct human
pose in indoor and outdoor datasets. Pavlakos \emph{et
 al.}~\cite{pavlakos2017harvesting} proposed a geometry-driven multi-view
approach that automatically annotated images with 3D poses starting from generic
2D detections~\cite{newell2016stacked}. Their harvested 3D poses are used to
demonstrate their effectiveness in two applications: 2D pose personalisation and
training a ConvNet from scratch for single view 3D human pose prediction.
Trumble~\cite{trumble2017total} made use of a CNN trained on probabilistic
visual hull data obtained from multi-viewpoint videos, and an LSTM framework to
exploit the temporal continuity of reconstructions.

Unlike  approaches such
as~\cite{elhayek2015efficient,pavlakos2017harvesting,trumble2017total}, we do
not perform pose estimation\footnote{Silhouettes in the
  case of \cite{trumble2017total}.} for each view before fusing them in a final
stage. Instead, we generalise  multi-stage approaches\cite{wei2016convolutional,tome2017lifting} to multiple
views, and iteratively seek an estimate consistent over all views. 

\section{Our Formulation}


We follow \cite{wei2016convolutional,tome2017lifting} in maintaining a
six stage algorithm. At each stage our CNN takes two inputs (see
Figure~\ref{fig:architecture_stage}): {\em (i)} the set of images from
different cameras we are trying to reconstruct from; and {\em (ii)}
the set of 2D pose heatmaps predicted in the previous stage for each
multi-view image. Inside each stage the algorithm independently
improves the 2D locations of joints in each image and uses them to
reconstruct a 3D model consistent with the 2D joint predictions for
all the views. Maintaining this internal representation of pose as a
3D model, coherent with all views, allows us to inject 3D information
into the learning process. In addition, by reprojecting the 3D model
into all the camera views using  known camera geometry we can use
2D losses throughout all the stages bypassing the need for 3D
annotations associated with the images.



This novel multi-view and multi-stage reconstruction allows us to {\em
  rethink joint locations} in light of knowledge of an interim 3D
reconstruction, to recover from mistakes made, and to try again to
find support in the image for the predictions of joint locations made
by a coherent working hypothesis of 3D positions. Details  are
given in section \ref{sec:network}.

 Importantly, our approach maintains the computable sub-gradients of Tome et
 al.~\cite{tome2017lifting} when generating and projecting the 3D model. This
 allows the system to be trained end-to-end. We make substantial changes that
 improve the robustness of the system while preserving the guarantees
 of~\cite{tome2017lifting} that the model fitting procedure will not get stuck
 in poor fitting local optima. This is done by replacing the Least Squares
 procedure of~\cite{tome2017lifting} with an Iterative Reweighted Least
 Squares ({\sc irls}) approach that mimics the Huber loss and preserves convexity
 for any particular choice of planar rotation. Details of this are given in
 section \ref{sec:3D-pose-estimation}.

\subsection{Details of the Network}
\label{sec:network}
Our proposed architecture is a multi-stage convolutional neural network inspired
by the work of Tome \etal~\cite{tome2017lifting}, which was in turn an extension
of the architecture introduced by Wei \etal\cite{wei2016convolutional}. They
proposed Convolutional Pose Machines (CPM), a multi-stage 2D pose estimator in
which each stage performed a refinement of the estimate computed
by the previous stage. 

As shown in Figure~\ref{fig:architecture_stage}, the first step in each
stage independently predicts, in every camera view, the 2D pose of
the person in the image. These predictions take the form of heatmaps generated
via a convolutional architecture with the weights shared between all camera views. 

These heatmaps are generated by a: {\em (a)} a set
of convolutional layers \emph{shared} by all stages that are performing feature
extraction; followed by {\em (b)} a set of convolutional layers, \emph{unique}
to each stage, that compute a heat map representing the location of each joint.
All stages (except stage 1) also take as input the heatmaps generated
in the previous stage.
The size and connections of these convolutional layers remain the same as in
CPM\cite{wei2016convolutional}. However, we additionally apply {batch normalization}
before the ReLu.

The next step within each stage takes heat-maps as input and computes
the 3D pose most consistent with the 2D information provided by each
camera view. Heat-maps are then converted into 2D locations by
selecting the most confident pixel as the location of each of the
joints
\begin{align*}
 I_p^c = \argmax_{(u,v)} H_p^c[u, v]
\end{align*}
where $H_p^c$ is the heat-map representing joint $p$ for camera view
$c$. These 2D poses are then used by the multi-camera probabilistic 3D
pose estimator (described in section~\ref{sec:3D-pose-estimation}) to
generate a single 3D pose that agrees over all the different camera 2D
poses. This pose is projected back onto the 2D image for each
camera view using a weak perspective projection, and the new projected
2D poses are converted into heat-maps by a Gaussian convolution 
\begin{align*}
 \hat H_p^c[u,v]=\begin{cases} 1 & \text{if} (u,v)=\hat{I}_p^c\\ 0 &
 \text{otherwise.}
 \end{cases}
\end{align*}
where $\hat{I}_p^c$ is joint $p$ of the projected 2D
pose in camera $c$.

The final operation fuses the heat-maps regressed by the convolutional
layers with those estimated by projecting the 3D pose into 2D. This
fusion is implemented by applying a convolutional layer with filters
of size $\left[1 \times 1 \right]$ and $\text{number\_joints}$
filters, to each camera view independently, giving a set of heatmaps,
one for each choice of  joint and camera.


As an implementation detail, all the computations performed on each
camera view make use of the same convolutional operations; this
enables us to have an efficient implementation by setting the batch
size to be equal to the number of cameras and ordering the images
appropriately.

 \subsection{Studio Setup and Camera Assumptions}
\label{sec:studio-setup-camera}
  We make use of the Human3.6M dataset~\cite{ionescu2014human3} for
  training and evaluation.  This dataset was generated in a
  multi-source capture studio with the ground-truth reconstructions
  coming from a ten camera Vicon studio, and four video cameras facing
  each another at right angles and far enough to fully capture a 4 by
  3 meter studio environment.

 Following the camera model and inference of~\cite{tome2017lifting}, we continue
 to assume a scaled orthographic model. Importantly, we assign the same choice
 of scale to all cameras. This assumption is noticeably stronger than the
 previous scaled orthographic reconstruction of~\cite{tome2017lifting}.
 With the four cameras facing
 towards each other, our stronger assumption does not allow increase in overall scale
 due to movements towards one camera, as this would correspond with movement away
 from another camera and a corresponding decrease in scale. However, it does
 allow for changes in scale of the object itself allowing our algorithm to
 handle people of different sizes.


 \subsection{Additional data}
 \label{sec:additional-data}
 One concern, when trying to show how additional data can lead to improved
 results in the 3D reconstruction of people, is the restrictive form of the
 Human3.6M evaluation dataset. With the limited appearance and repetitive
 range of actions, that occur both in the training and in the evaluation sets,
 networks trained on more general datasets might perform worse than those
 trained on restrictive datasets that are closer to the test data. To avoid such
 issues, we make use of an additional set of actors performing the same actions
 captured by the authors of the Human3.6M dataset.

As with many datasets  in computer vision,  Human3.6M was
originally subdivided into training, test and validation subsets; the
reconstructions for the test set were not made publicly available, to avoid
over-fitting. However, for historic reasons, the test set has gone largely
unused, with detailed evaluations being reported on the validation set. This
means that we have access to a publicly available additional corpus, composed of
unlabelled images from 2 men and 1 woman\footnote{Human3.6M dataset does not
  provide video for subject S10.}, captured in the same environment.

To illustrate how 3D data gathered by our method can improve existing results,
we augment two existing networks using this data. Our results show clear
improvement over published results, and help make the case not just that better
networks are needed for better results, but also more data.

Additional data can help 3D predictions in two separate ways, either
by improving the 2D localisation of joints, or by improving the 3D
lifting from the same 2D inputs. To show that our method returns
results of sufficiently high quality to improve both components, we
perform two separate experiments: (1) we show improvements on 2D joint
prediction while keeping the 3D lifting constant, and (2) we show how a
generic lifter that takes as input precomputed joint locations can be improved
by training on our additional 3D data.
\subsection{3D pose estimation}
 \label{sec:3D-pose-estimation}
 We now review the pose estimation of Tome et al.\cite{tome2017lifting} that
 generates a 3D pose from 2D joint locations; discuss its generalisation to
 multi-camera systems; and modifications to improve robustness to outliers and
 its stability.

 Tome \etal suggested approaching human pose estimation using a formulation
 inspired by non-rigid structure from motion. Assuming a known basis of human
 poses given by a set of matrices ${\bf e}${, and standard deviations $\sigma$,}  and a rest shape $\mu$, they suggest
 estimating the cost of a particular parameterised human pose, given 2D {locations} $I$, as:
 \begin{equation}\label{eq:get3D}
\argmin_{R,a}  ||I -s\Pi E R (\mu + a\cdot {\bf e} )||_2^2 + \sigma^2 \cdot a^2
 \end{equation}
 Where $\Pi$ is the canonical orthographic projection matrix, $E$ a known
 transformation from the world co-ordinates to those of the camera, $R$
 is a planar rotation matrix that describes the rotation of the human pose in the
 ground-plane, and $s$ is the estimated per-frame scale.
 Here $a$ is a vector of basis coefficients, ${\bf e}$ a 3D tensor of dimensions basis
 $\times$ points $\times$ 3.
 The tensor product $a\cdot {\bf e}$ is defined as $\sum_i a_i
 {\bf e}_i$, and the square terms in the final expression refer to an
 elementwise square. 
 The closest parameterised pose for 2D data $I$ was {given by minimising the  cost}:
 \begin{equation}
  \argmin_{s,a,R} P(s,a,R|I)
 \end{equation}
 The authors observed that, for any given choice of rotation, the global minima
 could be interpreted as an unconstrained linear least squares problem and
 solved efficiently. They suggested brute forcing over a small set of ground
 plane rotations to quickly find a global minima without needing to worry about
 getting stuck in poor quality local optima.

 We make several additions to this framework:
 \subsubsection{Rotation marginalisation for improved stability}
 Tome \etal ~\cite{tome2017lifting} observed that using more than $80$ rotations did not improve the
 overall accuracy of the reconstructions. Although this is true, their algorithm
 often yields flickering and unstable reconstructions when run on video data.
 Much of this flicker can be attributed towards trying to reconstruct ambiguous
 poses that can be equally well explained by two or more different rotations.
 We write the optimal reconstruction given a choice of rotation $R$ as \mbox{$\rec_R=R s(\mu + a \cdot {\bf e})$} where $a$, and $s$ are
 found by solving the following optimisation problem
 \begin{equation}
   \{s,a\}= \argmin_{s,a} P(s,a,R|I)
   \label{eq:minpose}
 \end{equation}
 Marginalising over the set of rotations $\cal R$, gives the following 3D body pose estimate:
 \begin{equation}
  \frac{\sum_{R\in \cal R}\exp(-\rho P(s_R,a_R,R|I))\rec_R}
  {\sum_{R\in \cal R}\exp(-\rho P(s_{R},a_{R},R|I))}
  \label{eq:average}
 \end{equation}
 
This elimination of flickering is highly desirable, not just in that it makes
the reconstructions of video appear more lifelike and appealing to humans, but
also in that the stability of the reconstructions carries important semantic
information. If we are to use 3D reconstructions of people as a first step in
action analysis, the stability and dynamics of the reconstructions contains
important information that informs our understanding of the actions.
\subsubsection{Principled shape warping for multiple views}
Tome \etal \cite{tome2017lifting}, approached the problem of reconstruction through the lens of
probabilistic PCA~\cite{Tipping99probabilisticprincipal} with a known basis. In
their framework, after generating a reconstruction from basis coefficients, a
final stage is to warp the reconstruction to lie closer to the input data. In
the context of 3D reconstruction from an single orthographic camera this can be
done as post processing, where a weighted average of the $x$ and $y$ coefficients
of the image and the reconstruction $\rec_R$ are taken together while the
$z$ component remains constant.

When multiple cameras are being used, this fusion between the model and the
data can not be performed as a simple post-processing step. Instead, we jointly
estimate a new shape $\tilde Q$ consistent with all frames and close to the
model estimate.
Given a rotation $R$, this can be written as
\begin{equation}
 \argmin_{\tilde\rec_R,s,a} \lambda\sum_{c\in \cal C}||I_C -\Pi E \tilde\rec_R ||_2^2 + ||\tilde\rec_R - s R (\mu + a \cdot {\bf e})||_2^2 +\sigma^2 \cdot a^2
 \label{eq:shape}
\end{equation}
where $\cal C$ refers to a set of cameras, $\lambda$ is a known scale
factor, and $E$ is the known external calibration that aligns world
co-ordinates with the camera's frame of reference.  As is standard in
geometry, this formulation finds the single body pose that best
explains all viewpoints; this is not equivalent to applying a single
camera approach to each view and averaging the results.
Again, this can be directly solved as an unconstrained least squares problem
given $R$; and as discussed in the previous subsection, we continue to
marginalise over the space of rotations.
\subsubsection{Robust losses for outlier rejection}
Finally, the use of the squared Frobenius norm as in the previous section makes the
reconstruction less robust to occlusions and to misdetected joints. If the
camera views were aligned, the first term of \eqref{eq:shape} would be minimised
by a pose that averages over the different predictions. Use of the Frobenius
norm would  mean
that if only one prediction is in the wrong place, it would ``pull'' the
reconstruction towards the mistake rather than discarding it as an outlier.
Instead we replace the squared Frobenius norm with a Huber loss.
\begin{equation}
 \argmin_{\tilde\rec_R,s,a} \lambda \sum_{c\in \cal C} ||I_c -\Pi E \tilde\rec_R ||_\epsilon + ||\tilde\rec_R - s R (\mu + a \cdot {\bf e})||_2^2 +\sigma^2 \cdot a^2
 \label{eq:shape2}
\end{equation}
where the Huber Loss $||x||_\epsilon=\sum_i |x_i|_\epsilon$ and 
\begin{equation}
 |x|_\epsilon =\begin{cases}
 \frac{|x_i|^2}{2} &\text{if $|x_i|\leq \epsilon$}\\
 \epsilon |x_i| - \frac{\epsilon^2}{2}&\text{ otherwise.}
 \end{cases}
\end{equation}
Although \eqref{eq:shape2} is not  a least square
problem, it can be solved as an iterative reweighted least squares problem ({\sc
irls}). In
practice, 5 iterations of least squares are sufficient to obtain a high quality
solution. Although robust to outliers, this new loss remains convex given a
choice of rotation, so local
minima are not a concern. The use of {\sc irls} for a fixed number of iterations
allows gradient propagation and end-to-end training as in \cite{tome2017lifting}.   

\section{Refining Existing Monocular Networks}
\begin{figure}[t]
  \includegraphics[width=\columnwidth]{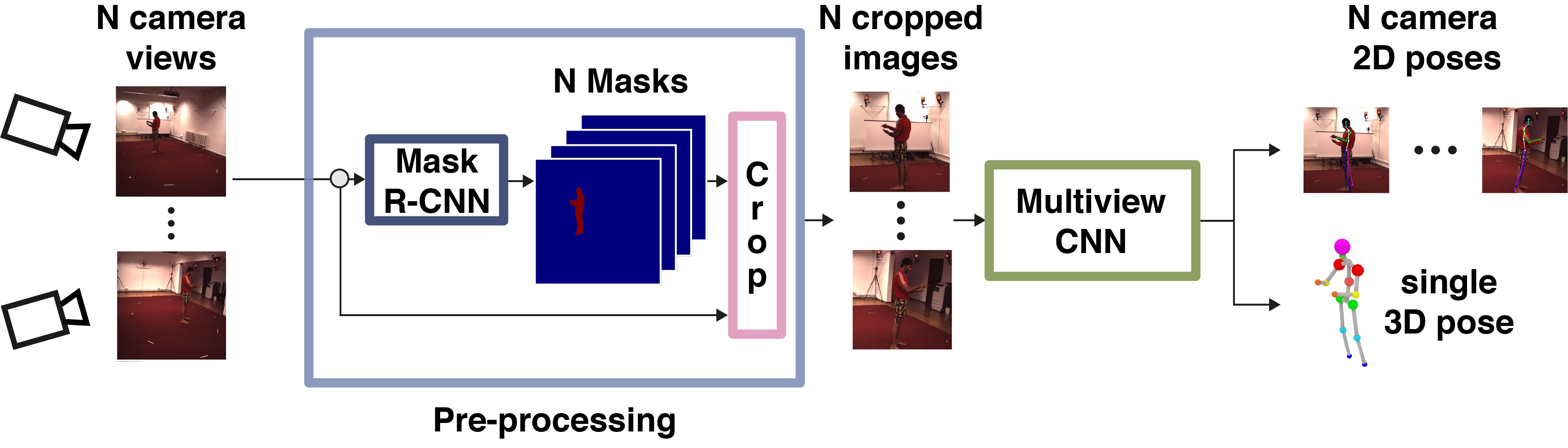}
  \caption{Labelling data using multi-camera 3D pose
   estimator.\label{fig:labelling}}
\end{figure}

Given the noticeable improvement in accuracy obtained by using
multiple cameras rather than just one (see table
\ref{tab:comparison_1}), it is natural to ask if our results can
improve the performance of existing networks by labelling
previously unlabelled data, and using this to augment the training
set. This labelling of new data can be seen in Figure
 \ref{fig:labelling}.  

Although conceptually simple, multiple small issues arise
from most experiments reporting results on an automatically
preprocessed version of the Human3.6M dataset.
   First, images are independently run through the
 \emph{Mask R-CNN} architecture \cite{he2017mask} in order to extract
 both the bounding box and the silhouette of the person represented in
 the images. This information is essential for cropping the area of
 the image containing the person in a similar manner to what is done
 on images with ground truth 2D data, guaranteeing that: 1) all the
 joints are inside the cropped region, centred around the hips; 2)
 the aspect ratio is one and 3) $25$ pixels of margin are added to the
 cropped region. These cropped regions are then used as inputs to our
 \emph{multi-camera network} which estimates 2D body poses for each 
 camera view and identifies the 3D pose most consistent with the set
 of 2D poses.  Finally, the 3D pose is projected into 2D for each 
 camera view using the known camera calibration.

 Data labelled by our approach is used to extend existing datasets. We
 simply treat the predicted bounding-boxes, 2D landmarks and 3D
 reconstructions the same way as  existing ground truth training
 data.
 
 \section{Experiments}
\label{sec:results}

\begin{table*}
  \hspace{-2mm}\resizebox{1.02\textwidth}{!}{
   \setlength\tabcolsep{2.0pt}
   \renewcommand{\arraystretch}{1.1}
   \begin{tabular}{lcccccccccccccccc}
    \toprule
    \textbf{Protocol 1}
    & \small Dir. & \small Disc. & \small Eat & \small Greet & \small Phone & \small Photo
    & \small Pose & \small Purch. & \small Sit & \small SitD. & \small Smoke & \small Wait
    & \small WalkD. & \small Walk & \small WalkT. & \small \textbf{Avg} \\
     \midrule
     \midrule
    LinKDE \cite{ionescu2014human3}  & \normalsize 132.7 & \normalsize 183.6 & \normalsize 132.4 & \normalsize 164.4 & \normalsize 162.1 & \normalsize 205.9 & \normalsize 150.6 & \normalsize 171.3 & \normalsize 151.6 & \normalsize 243.1 & \normalsize 162.1 & \normalsize 170.7 & \normalsize 177.1 & \normalsize 96.6 & \normalsize 127.9 & \normalsize 162.1\\
    Li \etal \cite{li2015maximum} & \normalsize - & \normalsize 136.9 & \normalsize 96.9 & \normalsize 124.7 & \normalsize - & \normalsize 168.7 & \normalsize - & \normalsize - & \normalsize - & \normalsize - & \normalsize - & \normalsize - & \normalsize 132.1 & \normalsize 69.9 & \normalsize - & \normalsize -\\
    Tekin \etal \cite{tekin2015predicting} & \normalsize 102.4 & \normalsize 158.5 & \normalsize 87.9 & \normalsize 126.8 & \normalsize 118.4 & \normalsize 185.1 & \normalsize 114.7 & \normalsize 107.6 & \normalsize 136.2 & \normalsize 205.7 & \normalsize 118.2 & \normalsize 146.7 & \normalsize 128.1 & \normalsize 65.9 & \normalsize 77.2 & \normalsize 125.3 \\
    Zhou \etal \cite{zhou2016sparseness} & \normalsize 87.4 & \normalsize 109.3 & \normalsize 87.1 & \normalsize 103.2 & \normalsize 116.2 & \normalsize 143.3 & \normalsize 106.9 & \normalsize 99.8 & \normalsize 124.5 & \normalsize 199.2 & \normalsize 107.4 & \normalsize 118.1 & \normalsize 114.2 & \normalsize 79.4 & \normalsize 97.7 & \normalsize 113.0\\
    Tome \etal \cite{tome2017lifting} & \normalsize 64.9 & \normalsize 73.5 & \normalsize 76.8 & \normalsize 86.4 & \normalsize 86.3 & \normalsize 110.7 & \normalsize 68.9 & \normalsize 74.8 & \normalsize 110.2 & \normalsize 172.9 & \normalsize 84.9 & \normalsize 85.8 & \normalsize 86.3 & \normalsize 71.4 & \normalsize 73.1 & \normalsize 88.4\\
    Pavlakos \etal \cite{pavlakos2017coarse} & \normalsize 67.4 & \normalsize 71.9 & \normalsize 66.7 & \normalsize 69.1 & \normalsize 72.0 & \normalsize 77.0 & \normalsize 65.0 & \normalsize 68.3 & \normalsize 83.7 & \normalsize 96.5 & \normalsize 71.7 & \normalsize 65.8 & \normalsize 74.9 & \normalsize 59.1 & \normalsize 63.2 & \normalsize 71.9\\
    Tekin \etal \cite{tekin2017learning} & \normalsize 53.9 & \normalsize 62.2 & \normalsize 61.5 & \normalsize 66.2 & \normalsize 80.1 & \normalsize 79.5 & \normalsize 64.6 & \normalsize 83.2 & \normalsize 70.9 & \normalsize 107.9 & \normalsize 70.4 & \normalsize 68.0 & \normalsize 77.8 & \normalsize 52.8 & \normalsize 63.1 & \normalsize 70.8\\
    Katircioglu \etal \cite{katircioglu2018learning} & \normalsize 54.9 & \normalsize 63.3 & \normalsize 57.3 & \normalsize 62.3 & \normalsize 70.3 & \normalsize 77.4 & \normalsize 56.7 & \normalsize 57.1 & \normalsize 79.0 & \normalsize 97.1 & \normalsize 64.3 & \normalsize 61.9 & \normalsize 67.1 & \normalsize 49.8 & \normalsize 62.3 & \normalsize 65.4\\
    Zhou \etal \cite{zhou2017towards} & \normalsize 54.8 & \normalsize 60.7 & \normalsize 58.2 & \normalsize 71.4 & \normalsize 62.0 & \normalsize 65.5 & \normalsize 53.8 & \normalsize 55.6 & \normalsize 75.2 & \normalsize 111.6 & \normalsize 64.15 & \normalsize 66.05 & \normalsize 51.4 & \normalsize 63.2 & \normalsize 55.3 & \normalsize 64.9\\
    Martinez \etal \cite{martinez2017simple} & \normalsize 51.8 & \normalsize 56.2 & \normalsize 58.1 & \normalsize 59.0 & \normalsize 69.5 & \normalsize 78.4 & \normalsize 55.2 & \normalsize 58.1 & \normalsize 74.0 & \normalsize 94.6 & \normalsize 62.3 & \normalsize 59.1 & \normalsize 65.1 & \normalsize 49.5 & \normalsize 52.4 & \normalsize 62.9\\
     \midrule
     \midrule
    Multi-View Martinez  & 46.5 & \bf 48.6 & 54.0 & 51.5 & 67.5 & 70.7 & 48.5 & 49.1 & 69.8 &\bf  79.4 & 57.8 & 53.1 & 56.7 & 42.2 & 45.4 & 57.0 \\
    PVH-TSP \cite{trumble2017total} & \normalsize 92.7 & \normalsize 85.9 & \normalsize 72.3 7 & \normalsize 93.2 & \normalsize 86.2 & \normalsize 101.2 & \normalsize 75.1 & \normalsize 78.0 & \normalsize 83.5 & \normalsize 94.8 & \normalsize 85.8 & \normalsize 82.0 & \normalsize 114.6 & \normalsize 94.9 & \normalsize 79.7 & \normalsize 87.3\\
    Pavlakos \etal \cite{pavlakos2017harvesting} & \normalsize \textbf{41.2} & \normalsize {49.2} & \normalsize 42.8 & \normalsize \textbf{43.4} & \normalsize 55.6 & \normalsize \textbf{46.9} & \normalsize \textbf{40.3} & \normalsize
    63.7 & \normalsize 97.6 & \normalsize 119.0 & \normalsize 52.1 & \normalsize \textbf{42.7} & \normalsize 51.9 & \normalsize \textbf{41.8} & \normalsize \textbf{39.4} & \normalsize 56.9\\
    \textbf{Ours} & \normalsize 43.3 & \normalsize 49.6 & \normalsize \textbf{42.0} & \normalsize 48.8 & \normalsize \textbf{51.1} & \normalsize 64.3 & \normalsize \textbf{40.3} & \normalsize \textbf{43.3} & \normalsize \textbf{66.0} & \normalsize 95.2 & \normalsize
              \textbf{50.2} & \normalsize 52.2 & \normalsize \textbf{51.1} & \normalsize 43.9 & \normalsize 45.3 & \normalsize \textbf{52.8}\\
    \hline
    \bottomrule
   \end{tabular}
  }
 \caption {Quantitative evaluation on the Human3.6M dataset. We report
   3D pose error results expressed in mm using the metric defined in
   protocol 1. All methods above the first line are monocular while
   those below (~\cite{trumble2017total,pavlakos2017harvesting} and
   {\bf Ours}) are multi-camera approaches.  `'Multi-View Martinez''
   refers to independently reconstructing from each monocular view
   using \cite{martinez2017simple} followed by averaging.
\label{tab:comparison_1}}
\end{table*}

We evaluate on Human3.6M using the two standard protocols for
evaluation. In \emph{protocol 1}, the training set consists of 5
subjects (S1, S5, S6, S7, S8), whereas the test set includes subjects
(S9, S11). The error metric is the Euclidean distance from the
estimated 3D joints to the ground truth, averaged over all 17 joints
of the Human3.6M skeletal model, and without alignment. The evaluation
is performed every $5^\text{th}$ frame, as in
\cite{zhou2016sparseness}, due to the high similarity of subsequent
frames.

\emph{Protocol 2}, introduced by \mbox{Bogo \etal \cite{bogo2016keep}}, uses
the same training and testing set as \emph{protocol 1}. However, evaluation
is performed on all frames captured by camera 3 during trial 1, and the error
metric reported is the average per-joint 3D error after aligning the
reconstruction with the ground-truth using Procrustes
analysis. 

Table \ref{tab:comparison_1} shows a comparison of our multi-camera
approach with other state-of-the-art techniques (both monocular and
multi-view) under \emph{protocol 1}. Our proposed approach outperforms
monocular methods, reducing the error by over $10$ milimetres, and
gives better results than the best multi-camera method of
Pavlakos~\etal~\cite{pavlakos2017harvesting} with an improvement of
more than $4$ milimetres. We also create a novel baseline based on
generating monocular reconstructions from each view using the method
of Martinez \etal \cite{martinez2017simple}, and averaging them after
alignment. This performs almost as well as Pavlakos~\etal, and is
reported in table \ref{tab:comparison_1} as ``Multi-view Martinez''.
Table \ref{tab:comparison_2} shows a comparison with other state of
the art approaches using \emph{protocol 2}.


\begin{table*}
  \hspace{-2mm}\resizebox{1.02\textwidth}{!}{
   \setlength\tabcolsep{1.9pt}
   \renewcommand{\arraystretch}{1.1}
   \begin{tabular}{lcccccccccccccccc}
    \toprule
    \textbf{Protocol 2}
    & \small Dir. & \small Disc. & \small Eat & \small Greet & \small Phone & \small Photo
    & \small Pose & \small Purch. & \small Sit & \small SitD. & \small Smoke & \small Wait
    & \small WalkD. & \small Walk & \small WalkT. & \small \textbf{Avg} \\
    \midrule
     Akhter \& Black \cite{akhter2015pose} 14j & \normalsize 199.2 & \normalsize 177.6 & \normalsize 161.8 & \normalsize 197.8 & \normalsize 176.2 & \normalsize 186.5 & \normalsize 195.4 & \normalsize 167.3 & \normalsize 160.7 & \normalsize 173.7 & \normalsize 177.8 & \normalsize 181.9 & \normalsize 176.2 & \normalsize 198.6 & \normalsize 192.7 & \normalsize 181.1\\
    Ramakrishna \etal \cite{ramakrishna2012reconstructing} 14j & \normalsize 137.4 & \normalsize 149.3 & \normalsize 141.6 & \normalsize 154.3 & \normalsize 157.7 & \normalsize 158.9 & \normalsize 141.8 & \normalsize 158.1 & \normalsize 168.6 & \normalsize 175.6 & \normalsize 160.4 & \normalsize 161.7 & \normalsize 150.0 & \normalsize 174.8 & \normalsize 150.2 & \normalsize 157.3\\
    Zhou \etal \cite{zhou2017sparse} 14j & \normalsize 99.7 & \normalsize 95.8 & \normalsize 87.9 & \normalsize 116.8 & \normalsize 108.3 & \normalsize 107.3 & \normalsize 93.5 & \normalsize 95.3 & \normalsize 109.1 & \normalsize 137.5 & \normalsize 106.0 & \normalsize 102.2 & \normalsize 106.5 & \normalsize 110.4 & \normalsize 115.2 & \normalsize 106.7\\
    Bogo \etal \cite{bogo2016keep} 14j & \normalsize 62.0 & \normalsize 60.2 & \normalsize 67.8 & \normalsize 76.5 & \normalsize 92.1 & \normalsize 77.0 & \normalsize 73.0 & \normalsize 75.3 & \normalsize 100.3 & \normalsize 137.3 & \normalsize 83.4 & \normalsize 77.3 & \normalsize 86.8 & \normalsize 79.7 & \normalsize 87.7 & \normalsize 82.3\\
    Tome \etal \cite{tome2017lifting} 14j & \normalsize - & \normalsize - & \normalsize - & \normalsize - & \normalsize - & \normalsize - & \normalsize - & \normalsize - & \normalsize - & \normalsize - & \normalsize - & \normalsize - & \normalsize - & \normalsize - & \normalsize - & \normalsize 79.6\\
     Moreno-Noguer \cite{moreno20173d} 14j & \normalsize 66.1 & \normalsize 61.7 & \normalsize 84.5 & \normalsize 73.7 & \normalsize 65.2 & \normalsize 67.2 & \normalsize 60.9 & \normalsize 67.3 & \normalsize 103.5 & \normalsize 74.6 & \normalsize 92.6 & \normalsize 69.6 & \normalsize 71.5 & \normalsize 78.0 & \normalsize 73.2 & \normalsize 74.0\\
     \midrule
     \textbf{Ours} 14j & \bf \normalsize 40.4 & \bf \normalsize 42.8 & \bf \normalsize 39.8 & \bf \normalsize 44.8 & \bf \normalsize 47.5 & \bf \normalsize 59.1 & \bf \normalsize 36.6 & \bf \normalsize 37.0 & \bf \normalsize 55.8 & \bf \normalsize 82.3 & \bf \normalsize 46.8 & \bf \normalsize 48.9 & \bf \normalsize  48.2 & \bf \normalsize  38.8 & \bf \normalsize 40.4 & \bf \normalsize 47.6\\
     \midrule
     \midrule
    Pavlakos \etal \cite{pavlakos2017coarse} 17j & \normalsize - & \normalsize - & \normalsize - & \normalsize - & \normalsize - & \normalsize - & \normalsize - & \normalsize - & \normalsize - & \normalsize - & \normalsize - & \normalsize - & \normalsize - & \normalsize - & \normalsize - & \normalsize 51.9\\
    Martinez \etal \cite{martinez2017simple} 17j & \normalsize 39.5 & \normalsize 43.2 & \normalsize 46.4 & \normalsize 47.0 & \normalsize 51.0 & \normalsize 56.0 & \normalsize 41.4 & \normalsize 40.6 & \normalsize 56.5 & \normalsize \textbf{69.4} & \normalsize 49.2 & \normalsize \textbf{45.0} & \normalsize 49.5 & \normalsize 38.0 & \normalsize 43.1 & \normalsize 47.7\\
    \midrule
    \textbf{Ours} 17j & \normalsize \textbf{38.2} & \normalsize \textbf{40.2} & \normalsize \textbf{38.8} & \normalsize \textbf{41.7} & \normalsize \textbf{44.5} & \normalsize \textbf{54.9} & \normalsize \textbf{34.8} & \normalsize \textbf{35.0}
                                  & \normalsize \textbf{52.9} & \normalsize 75.7 & \normalsize \textbf{43.3} & \normalsize 46.3 & \normalsize \textbf{44.7} & \normalsize \textbf{35.7} & \normalsize \textbf{37.5} & \normalsize \textbf{44.6}\\

    \bottomrule
   \end{tabular}
  }
 \caption {Quantitative evaluation of our approach against other methods using
  protocol  2 on 
  the Human3.6M dataset. Note that all other methods are monocular. The $14j$/$17j$
  annotation indicates the number of joints used in
  evaluation. \label{tab:comparison_2}}
\end{table*}

\begin{table}[tb]
  \resizebox{\columnwidth}{!}{
   \setlength\tabcolsep{6.0pt}
  \begin{tabular}{lcc}
   Formulation                      & Error \emph{Protocol 1} & Error \emph{Protocol 2} \\
   \toprule
   Squared Frobenius (no averaging) & 59.6 mm                 & 51.1 mm                 \\ 
   Squared Frobenius                & 59.4 mm                 & 51.8 mm                 \\
    Huber loss                      & 52.8 mm                 & 44.6 mm                 \\
    Huber loss (2 cameras)          & 64.2 $\pm$ 1.6 mm       & 52.8 $\pm$ 1.4 mm      \\
    \midrule
    GT Orthographic Triangulation   &27.9 mm                  & 20.7 mm\\
   \bottomrule
   \end{tabular}
  }
 \caption[width=\columnwidth] {Reconstruction error for different
   variants of our approach (see section \ref{sec:3D-pose-estimation}
   for details.) Huber loss (2 cameras) shows the mean and standard
   deviation of the reconstruction using only a pair of cameras at
   right angles with one another. GT Orthographic Triangulation shows
   the error due to the use of an orthographic camera, i.e. the the
   reconstruction error given perfect 
   detections.\label{tab:robustness_norms}}
\end{table}

Table~\ref{tab:robustness_norms} shows the importance of the changes to the pose
estimation made in~\ref{sec:3D-pose-estimation}; particularly the use of a more
robust Huber loss in place of the squared Frobenius norm, (Eq. \ref{eq:shape}
and Eq. \ref{eq:shape2}). 
Although, many works make use of the Huber loss as a more stable
approximation of the $\ell_1$ norm, this is not the case for us. Upon
inspection, we found that the optimal choice of $\epsilon$ that resulted in
the lowest 3D reconstruction error treated half of the joints with
$\ell_1$ norm and the other half with the squared Frobenius norm which
confirms that the Huber loss is effectively used to weigh the
relevance of each joint on a case by case basis.


A small improvement can also be seen from marginalising over the
rotations, although this modification primarily  improves the stability of
reconstructions rather than reducing the overall error. Finally we show how much
error can be attributed to the camera model, by
triangulating ground-truth detections under orthographic assumptions. This is
reported as ``GT Orthographic Triangulation''. 

\subsection{Improving Existing Monocular 3D Pose Networks}
Table \ref{tab:training_differences} shows the results of existing pose
estimation techniques~\cite{tome2017lifting,martinez2017simple} evaluated on a
variety of experiments where the models were trained using ground-truth
training data provided by the Human3.6M dataset \cite{ionescu2014human3}, and
additional unlabelled data (Subjects $\{S2, S3, S4\}$), automatically
labelled as previously described in section~\ref{sec:additional-data}.

In both approaches, we took the training hyper-parameters provided by the papers
and retrained the respective models using the augmented training data, without
fine-tuning the hyper-parameters.

The authors of~\cite{martinez2017simple} no longer have access to the retrained
stacked-hourglass 2D networks that they take as an input, so we can not
compute their 2D joint estimations on the held-out unlabelled data. Instead we
repeat their experiments, by training the network using the 2D poses estimated by
Tome~\etal~\cite{tome2017lifting} as input\footnote{The network of Tome \etal
  \cite{tome2017lifting} returns both 2D and 3D estimates of joint locations.},
and using these inputs to drive the 3D prediction. Without optimising the hyperparameters, this
leads to a noticeable decrease in the performance of the algorithm over that
reported by their paper, even though Tome~\etal report a lower 2D error than
that of  Martinez \etal. Despite this, we still observe a
substantial improvement in the 3D reconstruction from using more data. Note that
for this experiment, we do not update the 2D pose estimations, and all
improvement comes from the updated 3D estimator.

To illustrate that our method also improves  2D joint
localisation, we also retrain the network of Tome \etal. As an initial step in
training the algorithm, Tome \etal  compute a shape basis from MoCap data.
This basis is not updated during the end-to-end training of the pose estimator,
and the network itself is trained to improve 2D loss in joint predictions,
returning a 3D pose as a side-effect of its 2D pose computation. Although we
could update the 3D basis using our newly labelled data, we 
restrict ourselves to only updating the 2D pose predictor. As can be seen in
table~\ref{tab:training_differences}, this leads to a significant improvement in
2D error, and a corresponding reduction in the 3D error.


\begin{table}[tb]
    \renewcommand{\arraystretch}{1.1}
     \resizebox{\columnwidth}{!}{
 \setlength\tabcolsep{4.0pt}
 \begin{tabular}{ll|c|c|c|c}
 Approach & Experiment & \multicolumn{2}{c|}{Human3.6M dataset} & $\Delta$ & \% \\
 & & Train & Train + new data & & \\
 \hline
 Tome \etal             & 3D error (P\#1) & 88.4 mm & \textbf{84.4 mm} & 4.0 & 4.52 \\
 \cite{tome2017lifting} & 3D error (P\#2) & 70.7 mm & \textbf{67.2 mm} & 3.5 & 4.95 \\
                        & 2D error & 9.5 pix & \textbf{8.6 pix} & 0.9 & 9.47 \\
 \hline
 Martinez \etal           & 3D error (P\#1) & 75.8 mm & \textbf{72.5 mm} & 3.3 & 4.35 \\
 \cite{martinez2017simple}& 3D error (P\#2) & 57.6 mm & \textbf{55.9 mm} & 1.7 & 2.95 \\
 \end{tabular}
}
 \caption {Quantitative evaluation performed on existing approaches,
  demonstrating the performance gain when various models are trained with
  our additional data.  \label{tab:training_differences}}
\end{table}

\begin{figure}[tb]
 \hspace{-3mm}\includegraphics[width=1.05\columnwidth]{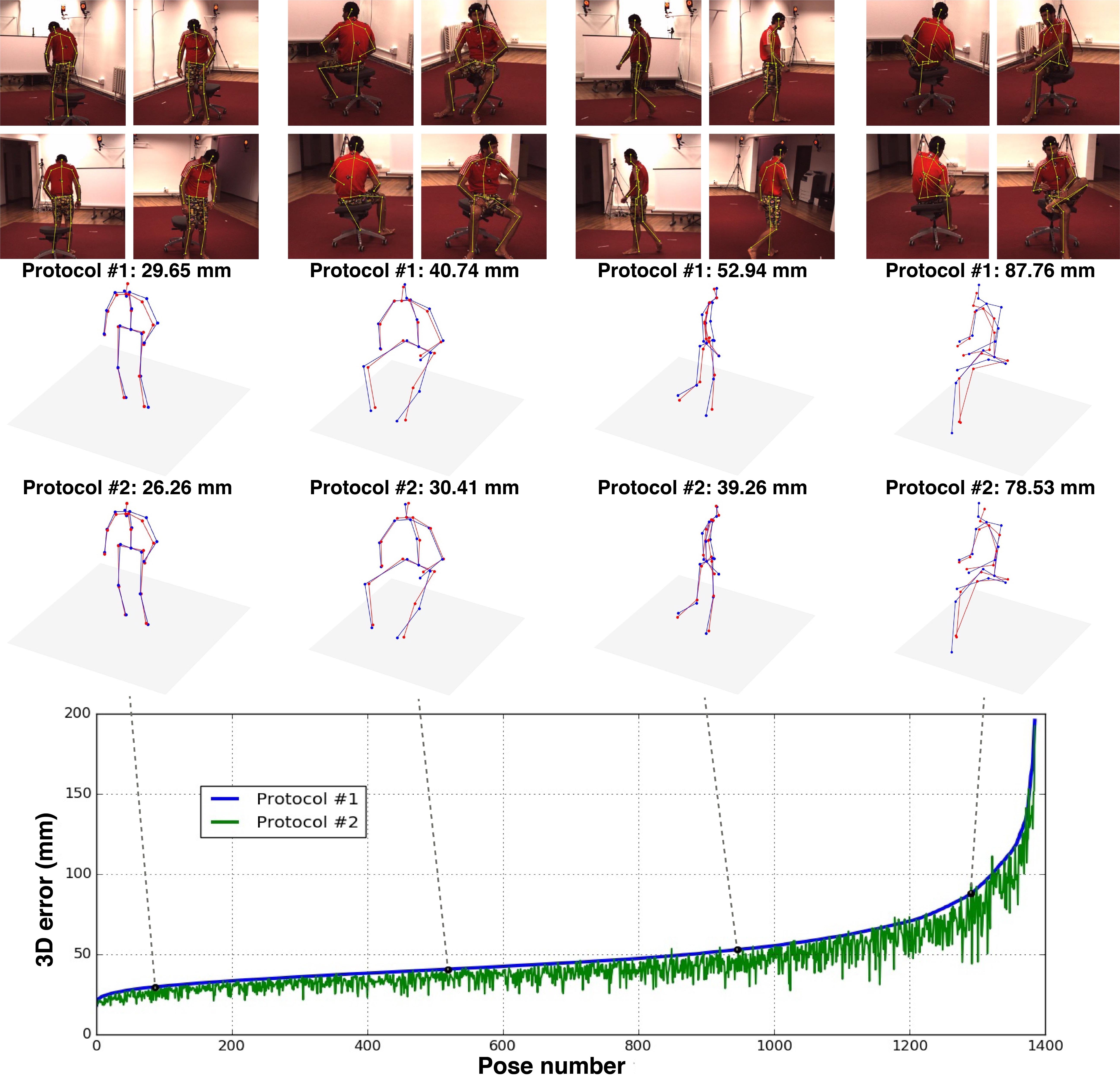}
 \caption{Multi-camera reconstructions showing sampled 3D errors from the test-set,
   sorted from small to large, for both protocol 1 and protocol
   2. Ground-truth reconstructions are given in blue, and the rows labelled
   protocol 1 and protocol 2 both show the same reconstructions in red, however
   protocol 1 shows the reconstruction {\em unaligned} with the ground-truth,
   and protocol 2 shows the reconstruction {\em aligned} to the
   ground-truth.  \label{fig:results} See section \ref{sec:results} for more
   details of the protocols.}
\end{figure}

Figure \ref{fig:results} shows some sampled 2D and 3D poses with the respective
reconstruction error for some multi-camera frames taken from the test-set of
Human3.6M dataset. The sorted error plot is based on sampling
the error every $10^\text{th}$ frame of trial 1.

\section{Conclusion}

We have shown a novel approach to markerless multi-camera 
motion-capture with a multi-stage architecture that allows us to
recover from initial misdetections, and still make use of image cues
in locating joints in subsequent stages.

We have demonstrated the clear benefits and robustness of our approach by
noticeably improving over existing multi-view markerless motion capture system.
In addition to this, we have shown how existing methods can be 
improved by using our approach as an initial first step to label otherwise
unlabelled data.

\small
\bibliographystyle{ieee}
\bibliography{egbib}
\end{document}